\documentclass[conference]{IEEEtran}
\IEEEoverridecommandlockouts
\usepackage[utf8]{inputenc}
\usepackage[T1]{fontenc}
\usepackage[hyphens]{url}
\usepackage{amsmath,amssymb,amsfonts}
\usepackage{tabularx}
\usepackage{algorithmic}
\usepackage{graphicx}
\usepackage{textcomp}
\usepackage{verbatim}
\usepackage{xcolor}
\usepackage{hhline}
\usepackage[all]{nowidow}

\usepackage[
pdftitle={Developing emotion recognition for video conference software to support people with autism},
pdfauthor = {Marc\ Franzen;\ Michael\ Stephan\ Gresser;\ Tobias\ Müller},
bookmarks=true,
bookmarksnumbered=true, 
bookmarksopen = true,
colorlinks = false]{hyperref}
\usepackage[style = ieee, urldate =comp]{biblatex}
\def\BibTeX{{\rm B\kern-.05em{\sc i\kern-.025em b}\kern-.08em
    T\kern-.1667em\lower.7ex\hbox{E}\kern-.125emX}}

\makeatletter
\newcommand{\linebreakand}{%
\end{@IEEEauthorhalign}
\hfill\mbox{}\par
\mbox{}\hfill\begin{@IEEEauthorhalign}
}
\makeatother

\begin{document}

\title{Developing emotion recognition for video conference software to support people with autism}

\def\arraystretch{1.5}

\author{
\IEEEauthorblockN{Marc Franzen}
\IEEEauthorblockA{\textit{Faculty E} \\
\textit{Ravensburg-Weingarten University}\\
88250 Weingarten, Deutschland}
\and
\IEEEauthorblockN{Michael Stephan Gresser}
\IEEEauthorblockA{\textit{Faculty E} \\
\textit{Ravensburg-Weingarten University}\\
88250 Weingarten, Deutschland}
\and
\IEEEauthorblockN{Tobias Müller}
\IEEEauthorblockA{\textit{Faculty E} \\
\textit{Ravensburg-Weingarten University}\\
88250 Weingarten, Deutschland}
\linebreakand
\IEEEauthorblockN{Prof. Dr. Sebastian Mauser}
\IEEEauthorblockA{\textit{Faculty E} \\
\textit{Ravensburg-Weingarten University}\\
88250 Weingarten, Deutschland \\
sebastian.mauser@rwu.de}
}

\maketitle

\begin{abstract}
We develop an emotion recognition software for the use with a video conference software for autistic individuals which are unable to recognize emotions properly. It can get an image out of the video stream, detect the emotion in it with the help of a neural network and display the prediction to the user. The network is trained on facial landmark features. The software is fully modular to support adaption to different video conference software, programming languages and implementations.
\end{abstract}


\section{Introduction}\label{sec:introduction}
Autism is a spectrum-condition, where the affected person can have a wide variety of impacts on various abilities. Some individuals have repetitive and stereotypical interests which makes them prefer doing recurring and monotonous work. Other individuals have impacts on social skills, like the ability to interact and communicate well with other people. \cite{Bobek:BA}\cite{auticon:autism}

Oftentimes, they lack the ability to detect emotions in the faces of their conversation partners. \cite{uljarevic2013recognition} This makes it difficult for them to estimate the course of an interview and therefore we want to support them with our software.

The software recognizes the emotion of a communication partner during a video conference and displays the result to the autistic individual. As a result, the software may compensate for a vital part of the social skillset of an autistic individual.
\section{Solutions}
In terms of emotion recognition of a facial image, there are already different commercial and non-commercial solutions with different accuracies available.
\subsection{Commercial solution}
In the commercial space, Affectiva is a company that develops emotion measurement technology and offers it as a service for other companies to use in their products. Their core product is the AFFDEX algorithm \cite{McDuff:2016:ASC:2851581.2890247}, that is used in Affectiva's AFFDEX SDK, which is available for purchase on their website \cite{affectiva-pricing}. It is mainly meant for market research, but it is also used in other environments, such as the automotive industry to monitor emotions and reactions of a driver. \cite{mcduff2014predicting}

AFFDEX has already found its way into a similar context as this project. With the help of AR glasses, children and adults can be assisted in learning "crucial social and cognitive skills" with a special focus on emotion recognition. \cite{affectiva-autism} This also supports our motivation, since many people with autism have problems understanding emotions.

Also, AFFDEX has a high accuracy in detecting "key emotions", more precisely in the "high 90th percentile". \cite{affectiva:accuracy}

As the AFFDEX SDK is a commercial solution, it is not easily accessible and has a price point of 25000 USD \cite{affectiva-pricing}. Therefore, this SDK is not the ideal solution for this research project and other solutions must be considered.

\subsection{Existing solutions}\label{sec:existing_solutions}
As an overview, we consult the paper \cite{cualeanu2013face}, which is summarized in the following paragraphs.
It distinguishes between three different steps:
\textit{Image Acquisition, Feature Extraction, Classification}.

\paragraph{Image Acquisition}

This step contains the acquisition of images from various sources, including "a database, a live video stream or other sources, in 2D or 3D". \cite{cualeanu2013face}

The data source can either be static by using still images or dynamic by using image sequences.

Afterwards, there might be pre-processing applied to the data. This could be de-noising, scaling and cropping to optimize the data for the next step.

\paragraph{Feature Extraction}

The extraction of features from facial data is an essential step, as they describe the "physical phenomena" \cite{cualeanu2013face} on which the detection of facial expressions is based on.

The better the selection of features and their representation of the face, the more robust the recognition afterwards is.

The available methods can be grouped into appearance features, geometric features, a hybrid approach using both and a template-based approach.

\paragraph{Classification}

In the final classification step, the detected facial expression is assigned to a predefined expression.

The available classifiers can typically be split into parametric and non-parametric machine-learning-methods, where either a predetermined function is given and parameters for this function are learned or the mapping function is learned without predetermining the form of a function.

Non-machine-learning methods might be feasible but are not discussed in this overview-paper.

\subsection{Examples}\label{sec:examples}

\paragraph{Linear Directional Patterns}
A solution referenced in \cite{cualeanu2013face} uses Linear Directional Patterns (LDP) representing the changes in the image as a histogram, which results in stable facial features.

These features are then fed into a Support Vector Machine (SVM) that maps these features to their corresponding emotions.

They state, that they reach an accuracy between 80\% and 99\% depending on the used parameters and test-scenarios.

\paragraph{Active Appearance Model}
Another already existing project uses an Active Appearance Model (AAM) to gather features which are then used to detect emotions in an image. \cite{ratliff2008emotion}
The AAM takes the statistical model of a face, representing its shape and tries to match this model to the image that is currently processed. More precisely, the model consists of a set of connected points (landmark coordinates) which are then iteratively deformed until they fit onto the current image. This process can be improved by training a model with images and their respective landmark coordinates. \cite{cootes1998active}

In that project, the extracted features are then used to calculate a mean parameter vector for each of seven emotions. These mean parameter vectors are then compared to the Euclidean distance from the face parameters of the current image. The accuracy of this approach is at around 90\% for the emotions fear, joy, disgust and neutral, but around 60-80\% for surprise, anger and sadness. \cite{ratliff2008emotion}

\subsection{Proposed solution}
Our hypothesis is that the direct matching from faces to emotions, as well as the manual feature engineering like in the above solutions with e.g. the Euclidean distance can lead to a lesser accurate detection than could be theoretically possible.

For similar tasks, neural networks are often considered, because of their ability to learn relevant correlations in the data on their own and it could perform better because of additional learned features.
  
Because of this, we propose the use of a neural network as our classifier to match faces to emotions, because they can learn features, a human would not consider to be relevant in the face.

To circumvent the problem to match raw image data to emotions, we take the common approach to use stable features similar to AAM in \cite{ratliff2008emotion} as the input to our neural network which then matches these to the corresponding emotions. This enables a more diverse prediction, without the need to distinguish between e.g. genders, skin-color and age.

\section{Method}
\begin{table}[htbp]
	\caption{Modules}
	\begin{center}
		\begin{tabular}{|c|p{.5\linewidth}|} 
			\hline
			\textbf{Module} & \textbf{Description} \\ 
			\hline
			\textit{main} & Starts the other modules and waits for them to exit. \\
			\hline
			\textit{input} & Gathers the image in some way and sends the image-data as .jpg over the socket. \\
			\hline
			\textit{model} & Receives the image-data and constructs a standardized OpenCV image object. \\
			\hline
			\textit{controller} & Receives the OpenCV image object, performs the emotion recognition and sends the conclusion as string. \\
			\hline
			\textit{view} & Receives the string and displays it on the screen. \\
			\hline
		\end{tabular}
		\label{tab:modules}
	\end{center}
\end{table}
Our solution is divided into four different kinds of modules (See Tab. \ref{tab:modules}).

Each module runs in a separate process and uses ZeroMQ for communication. ZeroMQ is a programming-language independent library, which enables us to open sockets on the localhost of a machine and then use TCP connections to exchange data \cite{zmq}. 
We implement this as a "Request-Reply" pattern, where one module requests data by sending "ready" over the socket, where the other module then sends the data in the previously defined format. If a module stops, it sends "done" and quits. As soon as another module receives "done" instead of "ready", it itself sends the "done" message to its adjacent module and quits. 

By implementing this pattern, we achieve an exchangeability, in which we can replace any module from any programming language, which supports OpenCV and ZeroMQ.

\subsection{Module: input}
We decide to use the static approach described in \ref{sec:existing_solutions} for better performance. For this, the input takes a screenshot using the mss library \cite{pypi:mss}, converts it to a .jpg format and sends it as a byte-stream to the model.
\subsection{Module: model}
The model expects any image type as a byte-stream and creates an OpenCV image object from that. To improve the performance, we scale this image down and thus reduce the amount of data for further processing. It is then serialized and forwarded to the controller.
\subsection{Module: controller}

\begin{figure}
	\centering
	\includegraphics[width=\linewidth]{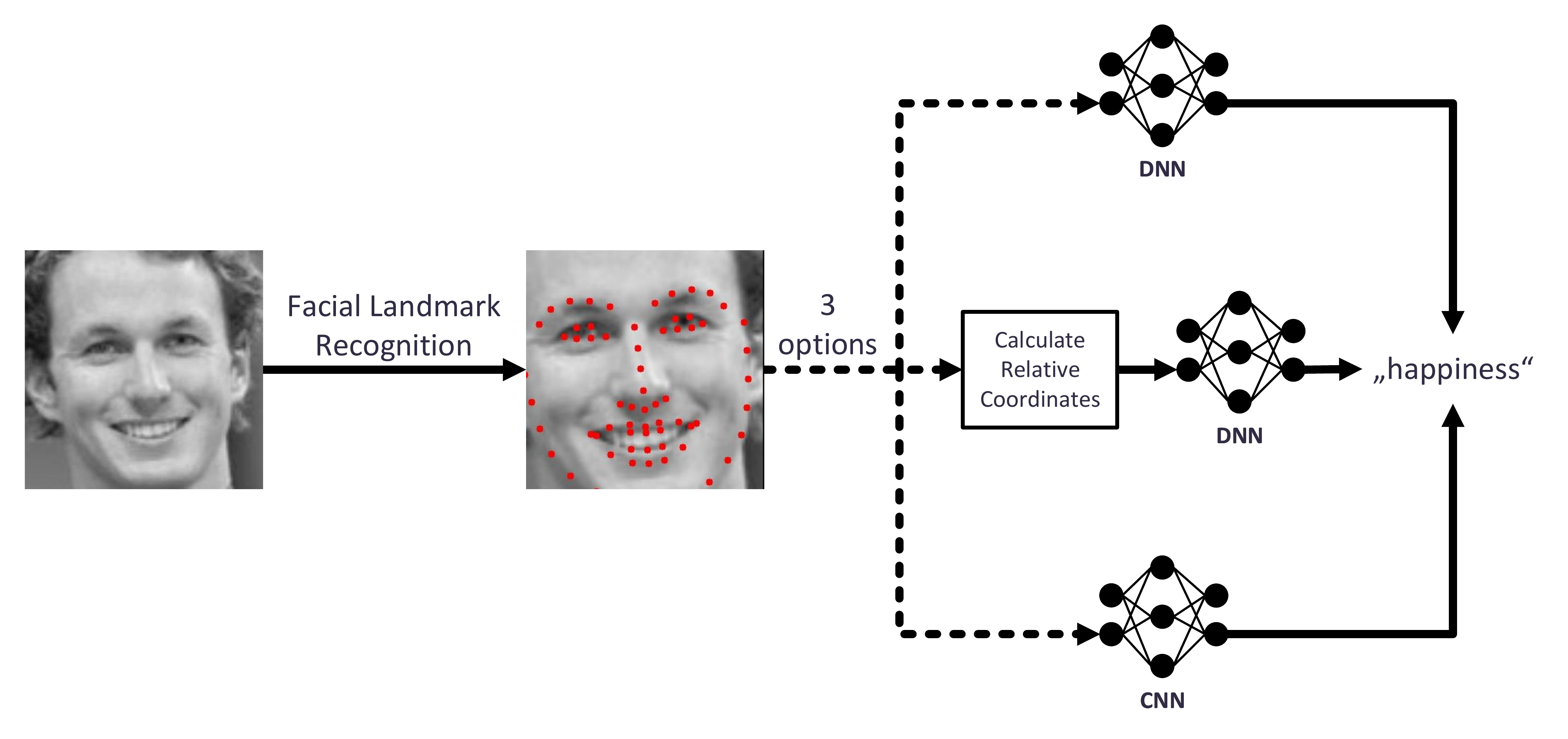}
	\caption{Steps for detecting emotions. Photo of face from dataset \cite{facialexpression:github}}
	\label{fig:emotiondetectionpipeline}
\end{figure}

The controller module expects a serialized OpenCV image from the model. It then de-serializes it and starts processing it in two steps, as shown in figure \ref{fig:emotiondetectionpipeline} and outlined in the following paragraphs.

\subsubsection{Facial landmark detection}
For facial landmark-detection we use the technique from \cite{kazemi2014one} that is included in the dlib library \cite{dlib} and adapted the implementation from \cite{rosebrock:faciallandmarks}, which is using the pre-trained detector inside the library. We choose this method from the summary \cite{Wu:CodeList} because it incorporates the face detection and the facial landmark detection in a single solution while providing real-time capabilities and an open source.

\subsubsection{Emotion detection}
From the gathered facial landmark points, we use artificial intelligence as a mapping function between the facial landmark features and the given emotion. The neural network learns the correlation between certain points of a face and the corresponding emotion while ignoring irrelevant points. As an example, humans would rate, that the position of the tip of the nose does not correspond as strongly to emotions as the corners of the mouth do.

\paragraph{Dataset}
We use the Dataset "facial\_expressions" \cite{facialexpression:github} from this challenge \cite{facialexpression:challenge}. This dataset has 13718 images, each labeled manually by the submitter with one of 8 facial expressions: anger, contempt, disgust, fear, happiness, sadness, surprise and neutral. We choose this dataset because it contains natural images of celebrities, as these reflect real emotions better than datasets crafted with actors.
For better training we remove the data submitted by "jhamski" and "628" as these images vary in resolution and color space and are sometimes distorted.
456 images were not included into the dataset because the chosen facial landmark recognition could not detect a face in them.
We also move random pictures of every category from the training dataset to a validation set to compare this solution against others (see Section \ref{sec:evaluation}). These images are not used to train the network.

The exact composition of the full possible training and validation sets is shown in table \ref{tab:dataset}.

Our final dataset consists of 12309 gray-scale images, each with a resolution of 350x350 pixels. The composition is unevenly distributed and thus not suitable for training neural networks. This is verified through initial tests, where all classes with small amounts of images are ignored in any predictions.

However, we determine, that the emotions contempt, disgust, fear and sadness are not important in a business video call. In most cases, happiness and neutral indicate if a conference is going well. Therefore anger and surprise are less important.

So, for the scope of this project we decide to leave them out for now and focus on happiness and neutral as facial expressions. If this works well, it should be expandable to other emotions with larger datasets.

This leaves us with 11228 images for training and 382 for validation.

\begin{table}[htbp]
	\caption{Number of images for each label}
	\begin{center}
		\begin{tabular}{|c|c|c|c|} 
			\hline
			\textbf{Emotion} & \textbf{Training} & \textbf{Validation} & \textbf{Importance}\\ 
			\hline
			Anger & 169 & 45 & medium\\
			\hline
			Contempt & 9 & - & low\\
			\hline
			Disgust & 12 & - & low\\
			\hline
			Fear & 11 & - & low\\
			\hline
			Happiness & 4961 & 191 & high\\
			\hline
			Neutral & 6267 & 191 & high\\
			\hline
			Sadness & 116 & - & low\\
			\hline
			Surprise & 288 & 49 & medium\\
			\hline
		\end{tabular}
		\label{tab:dataset}
	\end{center}
\end{table}

\paragraph{Training}
For training we use two common implementations of neural networks and some variation in the fed data which results in three total tested solutions.

We use TensorFlow \cite{tensorflow} with Keras \cite{Keras} as the framework to create all neural networks. 

\textit{Fully-Connected Neural Network:}

The first tested network is a dense neural network (DNN) to create a multi-layer perceptron (MLP). It has the following architecture:
\begin{itemize}
	\item Input Layer with X- \& Y-Coordinates for each of the 68 detected facial landmarks (136 inputs)
	\item Dense hidden Layer with 1024 neurons, followed by a dropout layer with rate 0.5
	\item Dense hidden Layer with 512 neurons, followed by a dropout layer with rate 0.5
	\item Dense hidden Layer with 256 neurons, followed by a dropout layer with rate 0.5
	\item Dense output layer with a node for each of the 2 possible emotions
\end{itemize}

As this model hits a plateau between 75\% and 80\% accuracy on the training and validation dataset, we suspect this could be caused by the changing absolute coordinates between images. So as a second attempt we add additional features and convert all absolute coordinates to relative ones.

\textit{Fully-Connected Neural Network with modified features:}

As the most relevant changes happen relative to the center of the respective portion of the face, the center point of each portion is calculated. This was added as an additional feature and all points of this portion were added relative to that center.

We decide to split the face in 4 parts: \textit{mouth, nose, left eye, right eye}. The eyebrows are given from the center of the corresponding eye, because we think that the movement of an eyebrow relative to the eye (raising eyebrows) is the most significant change in an emotion.

Additionally, the positions of the landmarks depend on the shape of a face. As an accommodation, we add the width, height and the center point of the face outline as features for the neural network. With these, the network can potentially predict the emotion independently from the geometry of the face.

This modified dataset uses the same DNN architecture with the only difference being the input layer to accommodate for the additional features.

This yields slightly better results on both datasets, as later discussed in \ref{sec:evaluation}.

\textit{Convolutional Neural Network:}

Convolutional Neural Networks (CNN) are usually used for image recognition, as they can make use of the spatial position of features. \cite{Wikipedia:CNN} So we think this concept might be applied here, since the position of each facial landmark can correspond to certain emotions.

For the dataset to be used with a CNN, all points of each image were projected into a 350x350 matrix, initially filled with zeros, where each added '1' corresponds to a facial landmark position.

The CNN has the following architecture:

\begin{itemize}
	\item Input layer with a 350x350 matrix with a depth of 1 for the single channel representing either a facial landmark at the given point or not
	\item 2D-Convolution layer with 32 neurons and a kernel size of 3
	\item Maximum 2D-Pooling layer with a size of 2x2 
	\item Dropout layer with a rate of 0.25 to combat overfitting
	\item Flattening layer to prepare the data for the fully connected layer
	\item Dense output layer with a node for each of the 2 possible emotions
\end{itemize}

To combat overfitting, additionally to the dropout layer, we use data augmentation since in this case it is easy to flip each matrix horizontally to get twice as many data rows as before.

\textit{Common parameters:}

All neural networks were trained using the optimizer Adam \cite{kingma2014adam} with a learning rate of $ 1*10^{-6} $.
The used activation function was always Rectified Linear Unit (relu), except for the output layer, which used softmax.
The range of the full width and height (350px each) of all coordinates for the DNNs was mapped to values from 0 to 1.

\textit{Model architecture creation:}

For a basis on which we can improve on, we use the model architectures provided by the Keras framework as examples \cite{Keras:Examples}.

With the Multilayer Perceptron and Convolutional Neural Network as starting points for our further development, we adapt them to our data format and reach the final presented networks by training, testing and altering the models.

The base remains conceptually the same, with minor modifications, like adding or removing layers and adjusting parameters, like the neuron-count or the dropout rate.

\paragraph{Use}
With the trained model, we run inference on the solution. The trained model is simply loaded using the Keras framework and inference is called using the given methods provided by the Keras Model class.

The result of this inference is then sent as a string for the emotion to the view module.

\subsection{Module: view}
The view accepts any kind of string and displays it in a window on the screen. In future work, there could be an implementation of this bachelor's thesis \cite{Bobek:BA}. 

\section{Results}\label{sec:evaluation}
To evaluate the software, we define the following office scenario: The opponent has a webcam with a resolution of 720p30 and the user has a PC with a recent (as of 2019, 9th Generation) Intel Core i5 processor. 

To compare our solution with a representative of the commercial sector, we selected the AFFDEX algorithm. It is implemented in the demo-program "AffdexMe" \cite{affectiva-affdexme}. We install it on the same machine we use for our own solution. Instead of the webcam input we use a stream of the desktop to simulate how AffdexMe would perform with the same input.

\subsection{Performance}
As AffdexMe does not provide interfaces for time measurement, we use a camera with a high framerate to count the milliseconds passed between the image being visible for AffdexMe and the program detecting the face and emotion. 

In our solution, we also measure the time between the image reaching the controller module and the detected emotion leaving the controller module for the different types of networks. 

All tests are executed on the same machine with an Intel Core i5 5500U and 8 GB of RAM.

The performance of each software varies greatly with the input resolution. For AffdexMe the input-images have a resolution of 525 x 525 pixels, while our software takes a screenshot with a resolution of 1366 x 768 pixels.

This has to be taken into consideration when evaluating the results in table \ref{tab:performance}. 

\begin{table}[htbp]
	\caption{Performance}
	\begin{center}
		\begin{tabular}{|c|c|c|} 
			\hhline{~|-|-|}
			\multicolumn{1}{c|}{} & \textbf{Average} & \textbf{Standard Deviation} \\ 
			\hline
			DNN & 737,1ms & 35,3ms \\
			\hline
			Modified DNN & 727,4ms & 25,5ms  \\
			\hline
			CNN & 767,8ms & 40,7ms  \\
			\hline
			AffdexMe & 367,8ms & 116,9ms \\
			\hline
		\end{tabular}
		\label{tab:performance}
	\end{center}
\end{table}

\subsection{Precision}
%

We compare all of our 3 optional solutions with each other and the solution from Affectiva \cite{McDuff:2016:ASC:2851581.2890247}. We separated random images from the dataset for validation, based on our rating of importance of each emotion for a business video conference.

\paragraph{Metrics}
We use two metrics to evaluate the results. For one, we collect the certainty for the emotions in AffdexMe and our solutions that correspond to the labeled emotions of the dataset and calculate the average per emotion. As Affectiva does not have a "neutral" emotion, we use the strongest detected emotion and calculate the reciprocal value. This is shown in table \ref{tab:certainty}.

\begin{table}[htbp]
	\caption{Certainty}
	\begin{center}
		\begin{tabular}{|c|c|c|c|} 
			\hhline{~|-|-|-|}
			\multicolumn{1}{c|}{} & \textbf{Happiness} & \textbf{Neutral} & \textbf{Total} \\ 
			\hline
			\textbf{Images tested} & 191 & 191 & 382 \\
			\hhline{|=|=|=|=|}
			DNN & 69,48\% & 75,04\% & 72,26\% \\
			\hline
			Modified DNN & 72,39\% & 76,79\% & 74,59\% \\
			\hline
			CNN & 55,27\% & 62,46\% & 58,87\% \\
			\hline
			Affectiva & 52,36\% & 26,54\% & 39,45\% \\
			\hline
		\end{tabular}
		\label{tab:certainty}
	\end{center}
\end{table}

As the other metric we calculate the accuracy of the predictions by each solution and AffdexMe. This is achieved by measuring the percentage of all correctly classified emotions on our dataset for validation. These results are shown in table \ref{tab:accuracy}.

\begin{table}[htbp]
	\caption{Accuracy}
	\begin{center}
		\begin{tabular}{|c|c|c|c|} 
			\hhline{~|-|-|-|}
			\multicolumn{1}{c|}{} & \textbf{Happiness} & \textbf{Neutral} & \textbf{Total} \\ 
			\hline
			\textbf{Images tested} & 191 & 191 & 382 \\
			\hhline{|=|=|=|=|}
			DNN & 74,35\% & 86,91\% & 80,63\% \\
			\hline
			Modified DNN & 78,01\% & 86,39\% & 82,20\% \\
			\hline
			CNN & 58,64\% & 75,39\% & 67,02\% \\
			\hline
			Affectiva & 59,26\% & 52,15\% & 55,70\% \\
			\hline
		\end{tabular}
		\label{tab:accuracy}
	\end{center}
\end{table}

\paragraph{History of accuracy while training}
\begin{figure}
	\centering
	\includegraphics[width=\linewidth]{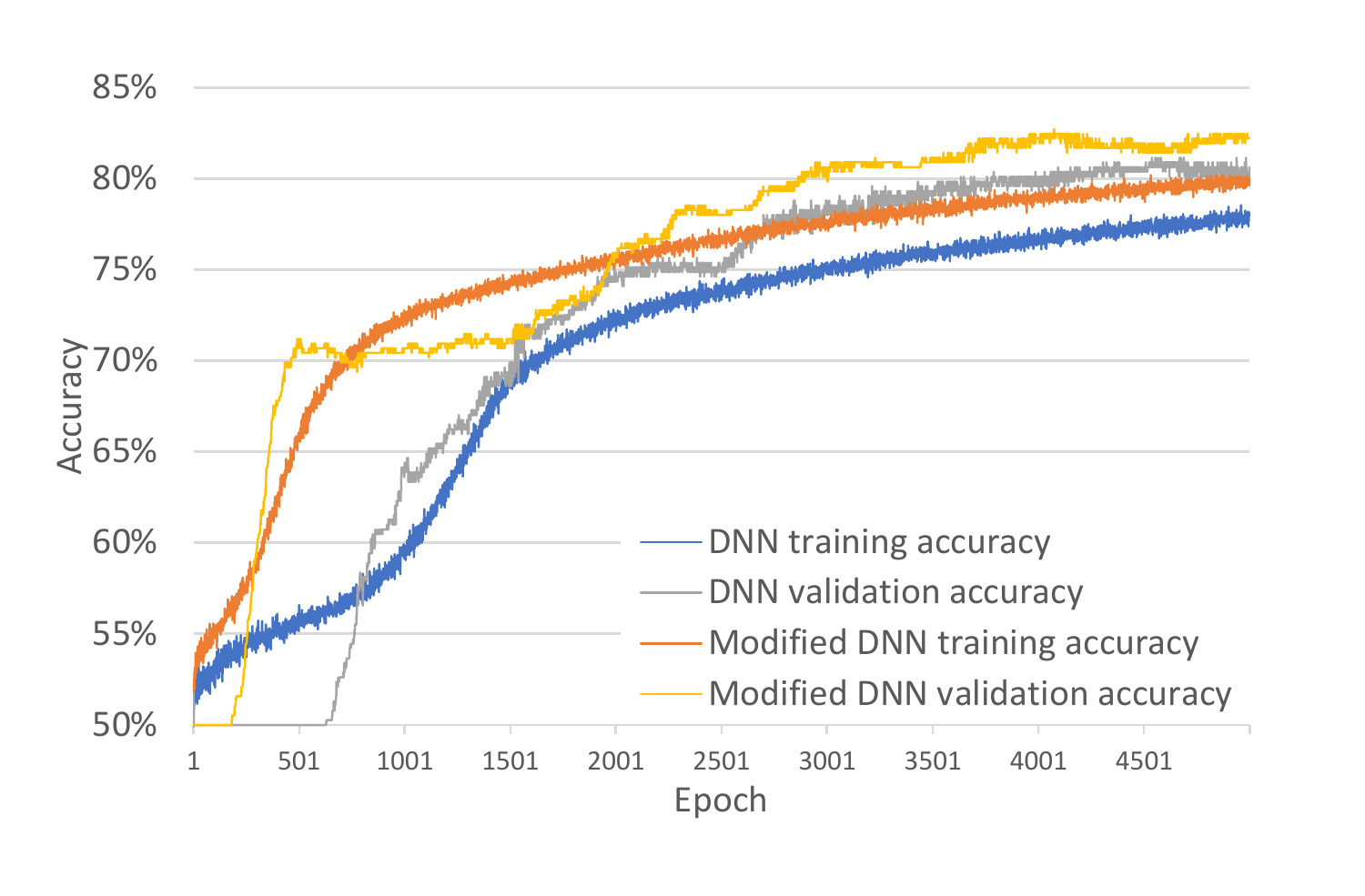}
	\caption{DNN: Accuracy on training and validation data over 5000 epochs}
	\label{fig:dnnaccuracyhistory}
\end{figure}

The DNNs are trained over 5000 epochs. The accuracy on both the training and validation datasets are shown in figure \ref{fig:dnnaccuracyhistory} over time.

Both versions increase steadily on the training dataset, while on the validation dataset it has more variation. The training accuracy increases faster with modified features, but they get to an equal growth towards the end.

With modified features it performs better overall, but only by a small margin. We suspect this is because the relative features have a slightly more relevant changes in their values and it might also be caused by the additional width and height features.

The training begins to plateau at about 3000 epochs. After that it only increases slowly. They might not be fully fitted yet, but are close to being fully trained, where it does not increase anymore or even decreases.

As the modified features train faster and get better overall results, this might be the preferred option when continuing to develop these methods further.

\begin{figure}
	\centering
	\includegraphics[width=\linewidth]{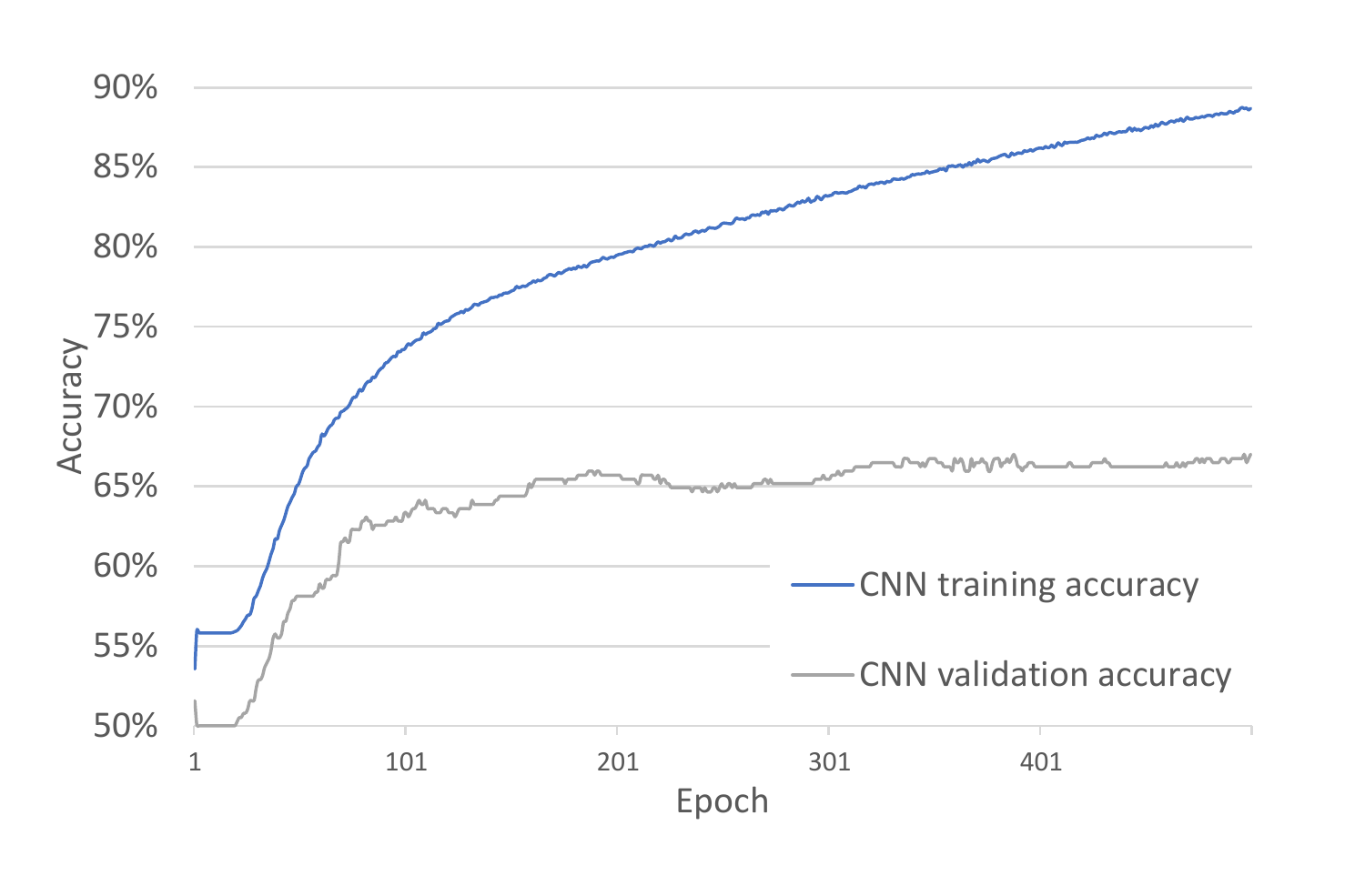}
	\caption{CNN: Accuracy on training and validation data over 500 epochs}
	\label{fig:cnnaccuracyhistory}
\end{figure}

Like the DNN, the CNN continues improving over the whole 500 trained epochs and also does not reach a point yet where any accuracy starts to decrease. From this we can also argue, that with more training time these results can be further improved. As training on the CNN takes 72 to 73 seconds per epoch, it is stopped at 500 epochs, which is equivalent to 10 hours of training time, because of the time constraint of this project.

We suspect it would get better quickly on the training dataset but would stay in the same range on the validation dataset, as it already plateaus early on and increases only by a small margin towards the end.

As it is shown in figure \ref{fig:cnnaccuracyhistory} the accuracy of the CNN is after 1/6th of the steps already more accurate on the training dataset. At the end of our training it is less accurate on the validation data than the DNN at its end.

To perform similar or better on the validation set than the DNNs, we think the CNNs architecture would need some modification and fine tuning.

\paragraph{Discussion}
As the DNN is smaller in file size of the model and is faster for inference on a single image, this is the preferred type for our solution for now, based on these results.

As already mentioned, we choose the DNN with relative feature points, as it is slightly better. The time for calculating relative coordinates from the absolute ones is negligible as the few needed array operations lay in the margin of error when measuring runtime.

Although the results show, that we do not reach accuracies on the validation-dataset in the higher 90th percentile, like the shown solutions in \ref{sec:examples}, we think that our solution has the benefit of using natural emotions for training.

We see many of the existing solutions use images of acted emotions, while our solution is trained on non-acted, everyday images of famous people.

To support this claim, we conducted several real-world tests with our solution in different conditions. They show a subjectively higher accuracy than our validation dataset.

Additionally, our solution cannot be easily compared with many other solutions, as we do not use the same datasets for validation. With different images, our accuracy can improve significantly when compared to the other solutions.

Finally, our solutions are more accurate on our chosen dataset than "AffdexMe", but also slightly slower. As already mentioned, this is due to the different resolutions.

Also, based on our experience using the AFFDEX algorithm, we think that it benefits from a stream of moving images. However, we use static images to test it to be able to compare the results to our solution.

Furthermore, our solution has the benefit compared to AFFDEX, that we are just training on the relevant emotions, while their solution is influenced by additional emotions which can result in more false predictions.
\section{Future Improvements}
\paragraph{Multi-modal information}
Because our social interaction is not just based on visual feedback alone, we propose the idea of combining several sources of human reactions as a result of the current emotional state.

This might improve the overall precision and confidence in the recognition system.

In case of video conference software, voice data might be taken into consideration. This was already evaluated in more detail in
\cite{chuang2004multi} and \cite{de1997facial}.

Additionally, chat messages can reflect the current emotional state and might also be used as another source of information, as discussed in \cite{wu2006emotion}.

These sources might be used as a starting point for further investigation.

\paragraph{Adaption to video conference software}
Because this project is built with modularity in mind, it is easy to provide adaptions for various video conference software and different User-Interfaces like the one presented in \cite{Bobek:BA}.

\paragraph{Larger Dataset}
Our dataset consists of 12309 images with labels. As the labels are distributed very unequally, future work can concentrate on gathering a larger and more evenly distributed dataset to improve the detection of the less important emotions as well as helping the classifier to distinct between similar emotions like anger and disgust.
\section{Conclusion}
We conclude that this software is a big step forward for the effective employment of individuals with autism.

Although our solution is limited to two distinct emotions, we think it can already help such individuals to achieve the everyday task of video conferences in an office environment.

And, as we stated before, with more data and further learning this can be improved to include more emotions.

It might be a good approach before deploying such a software, to discuss how this affects the emotional climate in the office, as well as what the effects of a wrong classification could be, if decisions are based on these judgements.

Overall, it is a helpful software that could already be used in its current state when all other hurdles about privacy, fairness and the possible impacts of misclassifications can be overcome.
\printbibliography

\end{document}